\documentclass[letterpaper, 10pt, conference]{ieeeconf}  

\IEEEoverridecommandlockouts                              

\usepackage{amsmath,amssymb,amsfonts}
\usepackage{graphicx}
\usepackage{hyperref}
\usepackage{booktabs}
\usepackage{tabularx}
\usepackage{subcaption}
\usepackage{siunitx}
\usepackage{makecell}

\title{\LARGE \bf
Focus on the Challenges: Analysis of a User-friendly Data Search Approach with CLIP in the Automotive Domain
}

\author{Philipp Rigoll$^{1}$, Patrick Petersen$^{1}$, Hanno Stage$^{1}$, Lennart Ries$^{1}$ and Eric Sax$^{2}$
\thanks{$^{1}$Philipp Rigoll, Patrick Petersen, Hanno Stage, and Lennart Ries are with FZI Research Center for Information Technology, 76131 Karlsruhe, Germany
        {\tt\small philipp.rigoll@fzi.de, petersen@fzi.de, stage@fzi.de, ries@fzi.de, sax@fzi.de}}
\thanks{$^{2}$Eric Sax is with Karlsruhe Institute of Technology, 76131 Karlsruhe, Germany
        {\tt\small eric.sax@kit.edu}}%
}

\begin{document}

\maketitle
\thispagestyle{empty}
\pagestyle{empty}

\begin{abstract}

Handling large amounts of data has become a key for developing automated driving systems.
Especially for developing highly automated driving functions, working with images has become increasingly challenging due to the sheer size of the required data.
Such data has to satisfy different requirements to be usable in machine learning-based approaches.
Thus, engineers need to fully understand their large image data sets for the development and test of machine learning algorithms.
However, current approaches lack automatability, are not generic and are limited in their expressiveness.
Hence, this paper aims to analyze a state-of-the-art text and image embedding neural network and guides through the application in the automotive domain.
This approach enables the search for similar images and the search based on a human understandable text-based description.
Our experiments show the automatability and generalizability of our proposed method for handling large data sets in the automotive domain.

\end{abstract}

\section{INTRODUCTION}
\begin{figure}%
	\centering
	\includegraphics[width=.98\columnwidth]{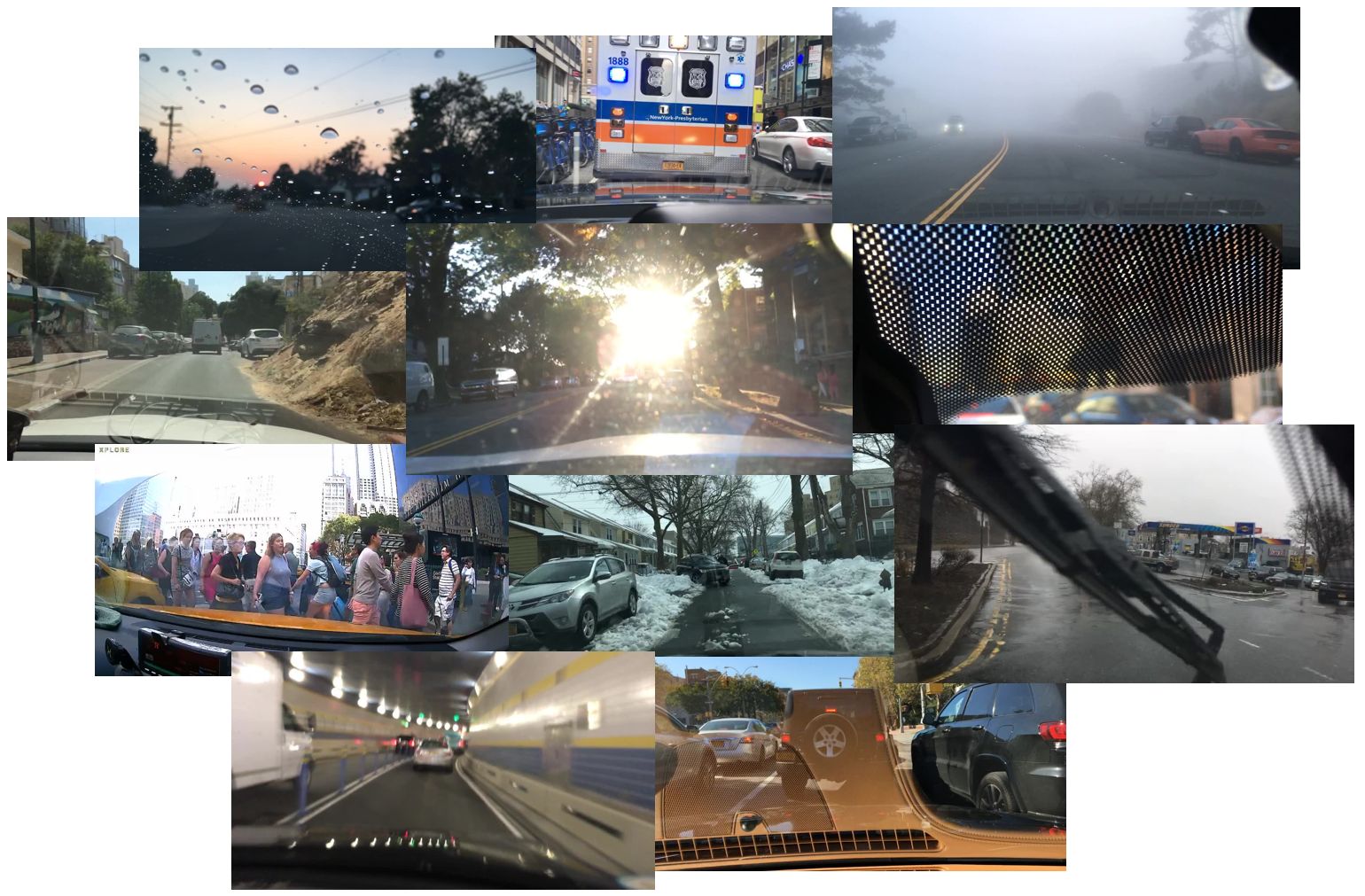}%
	\caption{Possible challenging images for an ADS found with the presented method in the BDD100k data set~\cite{yu_bdd100k_2020}.}%
	\label{fig:crazy}%
\end{figure}

Reliable data sets with their associated labels are key for the development of automated driving systems (ADS).
Camera images play an important role in the perception of ADS, therefore we focus on this data type.
Current state-of-the-art algorithms in computer vision benefit from increasing amount of data 
\cite{kuznetsova_open_2020}.
But as the data set sizes grow, manual annotation becomes costly \cite{chen_beat_2014,gevers_image_2003}.
However, many situations which challenge an ADS are identified during the development of the system.
Therefore, it is not possible to annotate these situations in advance.
In addition, the challenges could vary in their type (see Fig.~\ref{fig:crazy}).
For example, challenges could include a rare object class (e.g., carriages), special characteristics of an object (e.g., emergency car with flashing blue light), an environmental condition (e.g., sun dazzle), occlusions (e.g., a windshield wiper) or an image property (e.g., overexposure).

To utilize the presence of these situations in a data set, it is necessary to have a method to identify them in a pile of highly redundant data.
However, applying manual rule-based image search is not feasible due to the sheer amount of assumptions which need to be made.
But it is also inconvenient to train a separate neural network for the identification of every situation or every object, because one would then have to manually arrange separate training datasets.

Furthermore, for the training of neural networks, it is necessary to have enough labeled training data.
This is especially true for situations and objects that rarely occur.
Since these rare cases, with only a few examples in the data set, are only seen seldom during training, they can be a challenge for a neural network.
Our goal is to identify these cases with a generic method which is not dependent on labels and which is adaptable to find diverse situations and objects.
In this way, the manual, time-consuming labeling can be limited to fewer situations.
And it is also ensured that the identified, rare situations are taken into account.
In addition, our method should be intuitive enough to allow an ADS developer to focus on the development of their system.

The paper is organized as follows: Section~\ref{sec:related_work} gives an overview of the state of the art in image data handling. 
In Section~\ref{sec:method} we introduce our method for textual and image-based search.
Section~\ref{sec:evaluation} covers the performance of the proposed procedures on automotive data sets.
Finally, Section~\ref{sec:conclusion} concludes the paper and discusses future work based on the results.

\section{RELATED WORK}
\label{sec:related_work}
State-of-the-art data sets for the development of ADS consist of millions of frames and are increasing in size~\cite{bogdoll_ad-datasets_2022}.
Therefore, data handling becomes a key discipline as part of Automotive Systems Engineering~\cite{petersen_towards_2022,fingscheidt_analysis_2022-1}.
More precisely, it is necessary to identify certain data sets to be able to use them specifically in the development of ADS.
In particular, it is important to study the performance of an ADS in specific situations~\cite{li_coda_2022}.
To identify these situations, one can utilize metadata~\cite{qi_anomaly_2018}.
In our previous work, we showed that metadata based methods can be used to filter by geographical information and sun positions~\cite{rigoll_scalable_2022}.
This type of approach is independent of the image data itself.
Therefore, the method is fast and scalable to large data sets.
But at the same time, this approach depends on the availability of suitable metadata.
Since this approach is not very generic, the appropriate metadata must be aggregated for each situation and an appropriate filter routine must be programmed.
Lv et al.~\cite{lv_image_2004} present a method which allows the search for similar images by a compact feature vector.
However, the method only allows images as search queries and is therefore limited in application.
A comprehensive overview of interactive image search is given by Thomee et al.~\cite{thomee_interactive_2012}.
The presented approaches find images via an interactive dialog with the user.
However, the number of required training data is identified as a challenge.
In contrast, we investigate a method that is pre-trained and does not require any additional data specifically labeled for the task.

To offer a more user-friendly data handling method, we focus on human understandable text-based descriptions.
Hence, a method is needed which can establish a connection between images and textual descriptions.
To achieve this, Wang et al.~\cite{wang_annosearch_2006} started with a query word for an image and got a textual description by searching the web semantically and with visually similar images.
Alkhawlani et al.~\cite{alkhawlani_text-based_2015} reviewed different types of content-based image retrieval methods.
They examined text-based approaches which are comparable with the already discussed metadata method.
But they also remark the limited automatability of the annotation process and the fact that it is hard to describe all aspects of an image with a short textual description.
There are methods to caption images automatically~\cite{hossain_comprehensive_2019}.
But the problem with the limited expressiveness of text captions remains.
This is especially challenging when captioning automotive images, where the same objects, such as the road and cars, are almost always the dominant image components.
To achieve a more comprehensive description of images, a different representation is required.
Radford et al.~\cite{radford_learning_2021} introduce a method to transfer images and texts into a common latent space.
Their method is called CLIP (Contrastive Language Image Pre-training) and trained on a large data set with about 400~million (text, image) pairs.
CLIP provides a vector embedding for an input image or an input text.
During training, it is ensured that similar texts and images lead to similar vectors.
This way, CLIP solves a wide set of tasks like object detection, optical character recognition, geo-localization, action recognition, without the need of additional training and task-specific training data.
The similarity of the vectors is calculated utilizing the cosine similarity~\cite{national_institute_of_standards_and_technology_nist_cosine_2023}.
We examine whether and how CLIP can be used to solve the data handling challenges we identified.

\section{METHOD}
\label{sec:method}
To effectively support the development of ADS, we have the following requirements on our method:
\begin{itemize}
	\item generalizability to diverse situations and objects,
	\item short runtime for the sensible use as an assistive tool,
	\item scalability to large data sets,
	\item user-friendliness without the need of knowledge about the method.
\end{itemize}

Our approach comes into play when the ADS should be challenged with specific situations.
This can be of interest when the ADS shows peculiar behavior or low performance in specific situations, for example, when a rarely observed object appears.
Another reason may be that due to the known functional weaknesses of the ADS, situations are known that may present challenges.
An example of this is sunlight shining directly into the camera.
The methods presented in the following allow the search for corresponding situations in the data set and enable targeted tests.

\begin{figure*}%
	\centering
	\includegraphics[width=.9\textwidth]{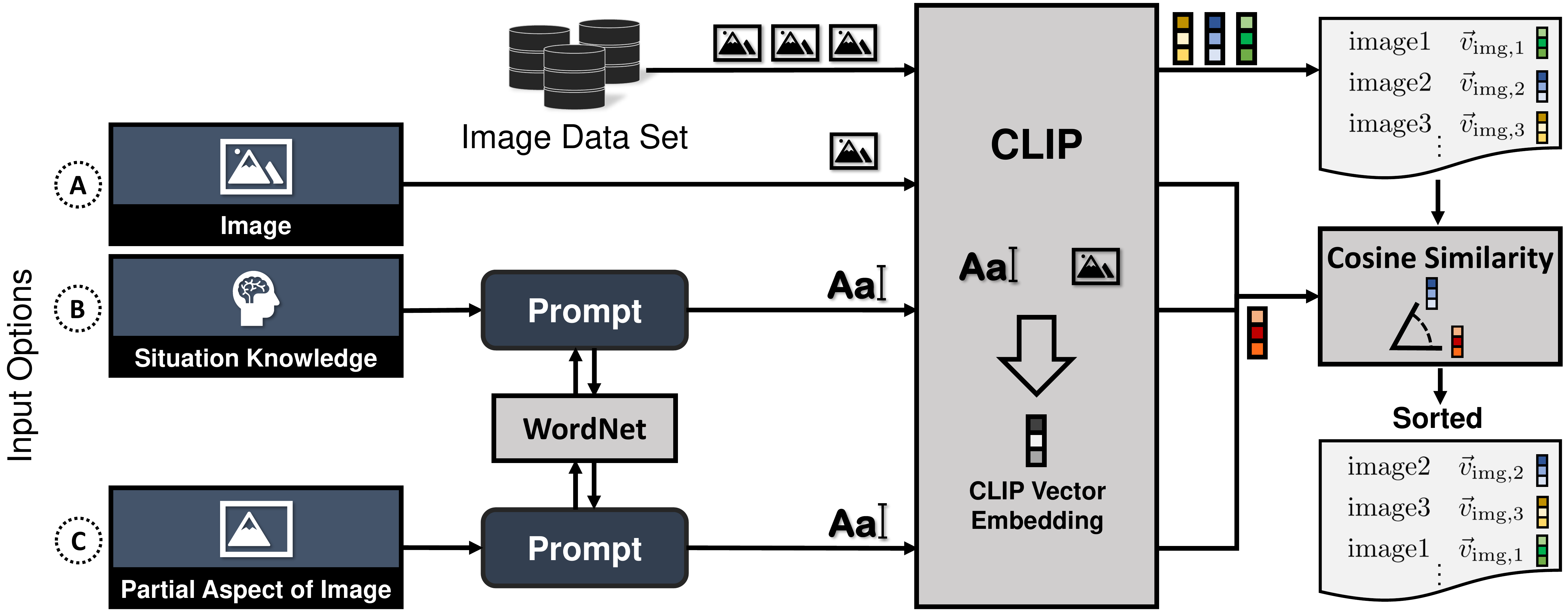}%
	\caption{Overview of the proposed method for textual and image-based search of similar images.}
	\label{fig:method}%
\end{figure*}

Preparatory for every image in the data set, the CLIP vector embedding $\vec{v}_{\text{img}}$ is calculated.
This calculation must be performed only once, and the resulting vectors can be saved in a vector database to increase the speed of the following operations \cite{wang_milvus_2021,johnson_billion-scale_2017}.
Vector databases were developed to handle high-dimensional vector data.
They speed up the calculation of vector operations, such as the similarity determination between vectors.
The next steps depend on how the corresponding situation can be described (see Fig.~\ref{fig:method}).
In the following, the description is called prompt and can be a text or an image.

\subsection{Specific image}
If the ADS shows peculiar behavior when exposed to a specific image, this image can be used as the prompt image to find similar situations.
To start the search, the CLIP vector embedding $\vec{v}_\text{sample}$ of this image must be calculated and can be used as a prompt to search for similar images.

\subsection{Situation knowledge}
If functional weaknesses of the ADS are known, the behavior should be investigated in similar situations to get a broader understanding of the weaknesses.
In the case that no corresponding situations are known in the data set and the search with a corresponding image is not sufficient, a different method can be used.
Therefore, the required situations are described textual, and these descriptions are then used as prompts.

It is also possible to describe the opposite of what is searched for and use this description as a negative prompt.
With negative prompts, images are found, which are dissimilar to the prompt.

\subsection{Partial aspects of an image}
Textual descriptions can also be useful, if only a partial aspect of an image has influence on the peculiar behavior of the ADS.
In this case, the search for similar images as a whole will not lead to satisfactory results.
Therefore, one has to develop textual descriptions of the crucial object or the partial aspect of the image.

To refine the textual description and create a precise prompt, related words are searched.
To systemize and partly automate this prompt generation, a lexical database like WordNet~\cite{miller_wordnet_1995} can be utilized.
This way, synonyms for a term, such as an object class, can be found.
Besides synonyms, hypernyms and its opposite hyponyms can be identified.
Hypernyms are more generic terms which subsume a set of terms.
Additionally, meronymy and holonymy can be used to name components or a whole.
If antonyms are identified, they can serve as negative prompts (as an example, see Table~\ref{tab:wordnet}).
\newline

In all cases, one will end up with a prompt, which could be an image or a string.
Next, the CLIP vector embedding $\vec{v}_\text{prompt}$ of the resulting prompt is calculated.
To sort the images $i$ by their similarity to the prompt, the cosine similarity between each image embedding and the prompt embedding is calculated: $s_i = \text{sim}_\text{cos}(\vec{v}_{\text{img},i}, \vec{v}_\text{prompt})$.
Where the cosine similarity \cite{national_institute_of_standards_and_technology_nist_cosine_2023} between the $n$ dimensional vectors $\vec{x}$ and $\vec{y}$ is defined by:
\begin{equation}
	\text{sim}_\text{cos}(\vec{x}, \vec{y}) = \frac{\sum^n_{i=1} x_i y_i}{\sqrt{\sum^n_{i=1}x_i^2} \sqrt{\sum^n_{i=1}y_i^2}}.
\end{equation}
To speed up this process, the similarity calculation can be parallelized.
The results are used as sorting keys, where the highest value indicates the most similarity between prompt and image.
Negative prompts are treated as normal prompts and also lead to a similarity score $s_{\text{neg},i}$.
Therefore, the final similarity measure is given by $s_{\text{result},i} = s_{\text{pos},i} - s_{\text{neg},i}$, where $s_{\text{pos},i}$ is the initial positive prompt.

\section{EVALUATION}
\label{sec:evaluation}
To investigate the performance of the proposed procedures on automotive data sets, we conduct some exemplary experiments, evaluate the performance quantitatively on a relevant real-world classification problem and compare the results qualitatively to existing methods.

\begin{figure}%
	\centering
	\includegraphics[width=.98\columnwidth]{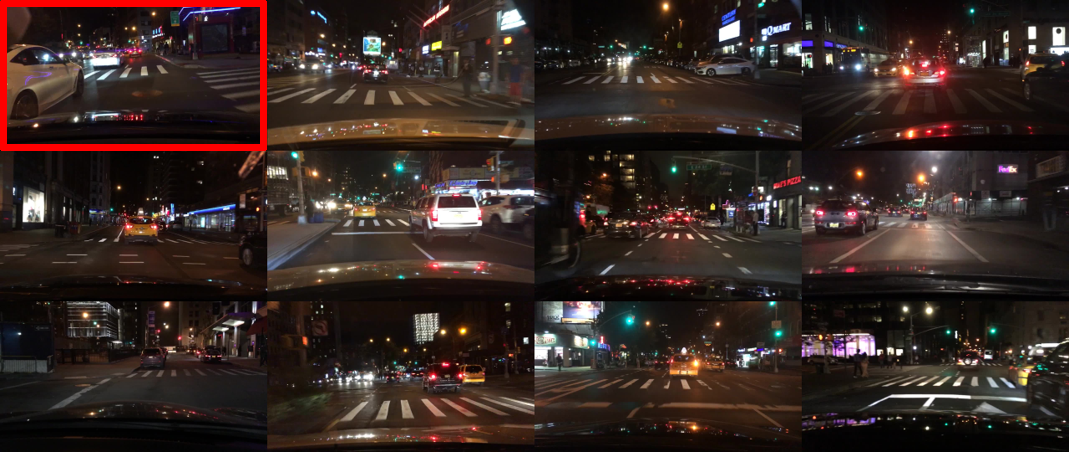}%
	\caption{The query image is in the upper-left corner in a red frame. The most similar images in the BDD100k data set~\cite{yu_bdd100k_2020} are sorted from left to right and top to bottom.}%
	\label{fig:similar}%
\end{figure}

\begin{figure}%
	\centering
	\includegraphics[width=.98\columnwidth]{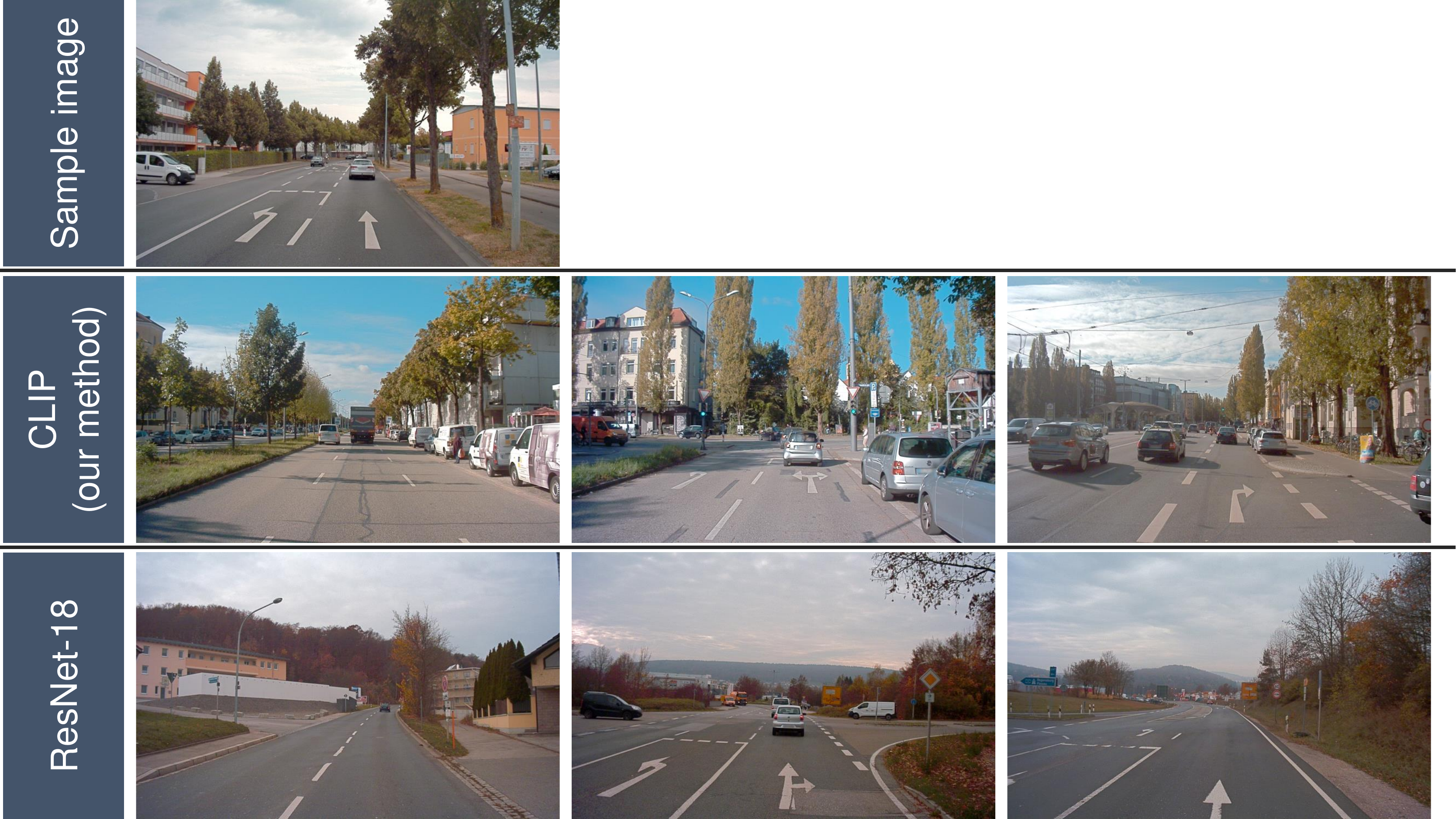}
	\caption{Comparison of the most similar images found in the A2D2 data set~\cite{geyer_a2d2_2020} by using our proposed method (CLIP) and ResNet-18.}%
	\label{fig:resnet_vs_clip}%
\end{figure}

In advance, the CLIP embedding for the whole data set is calculated.
The introductory exemplary experiments are performed on the BDD100k~\cite{yu_bdd100k_2020} and A2D2 data sets~\cite{geyer_a2d2_2020}.
First, we examined if similar images can be found as proposed.
Therefore, we randomly selected an image from the BDD100k data set.
The corresponding CLIP embedding is then used to calculate the cosine similarity for every image in the dataset.
After sorting by this similarity score, the most similar images can be identified.
In Fig.~\ref{fig:similar} the query image is marked red in the top-left corner.
The example image is taken at nighttime, a zebra crossing is visible and there are a few cars around.
The most similar images found by the proposed method also contain these elements.
Previously, this task could be performed by extracting the results of the intermediate layer of a convolutional neural network trained on a different task~\cite{razavian_cnn_2014}.
This intermediate feature vector was then used as an embedding. 
We want to compare CLIP to this method, therefore we take a look at the similar images found by CLIP and by the classical approach.
Thus, the output of the penultimate layer of a ResNet-18~\cite{he_deep_2016} (pretrained on ImageNet~\cite{deng_imagenet_2009}) is used to vectorize the images.
When we compare the result of CLIP and ResNet-18 (see Fig.~\ref{fig:resnet_vs_clip}), we find clues that  ResNet-18 seems to focus a bit more on the pixel level where CLIP is able to abstract to the semantic level.
In the example CLIP identifies the avenue in the image, and thus the found similar images also show typical properties of an avenue (e.g., trees on both sides of the road, within cities).
The images found by CLIP cover different weather and light conditions and contain many cars.
In contrast, ResNet-18 finds images which are missing properties of a typical avenue and two out of three contain fewer cars.
However, the images found by ResNet-18 are similar regarding their weather and light conditions.
Another hint that ResNet-18 is focusing on the pixel level is given by lane marking arrows in the second result image, which are at a very similar position as in the sample image.
However, CLIP identifies that there should be lane marking arrows on the street, but does not preserve the correct position in the image.

\begin{figure}%
	\centering
	\includegraphics[width=.98\columnwidth]{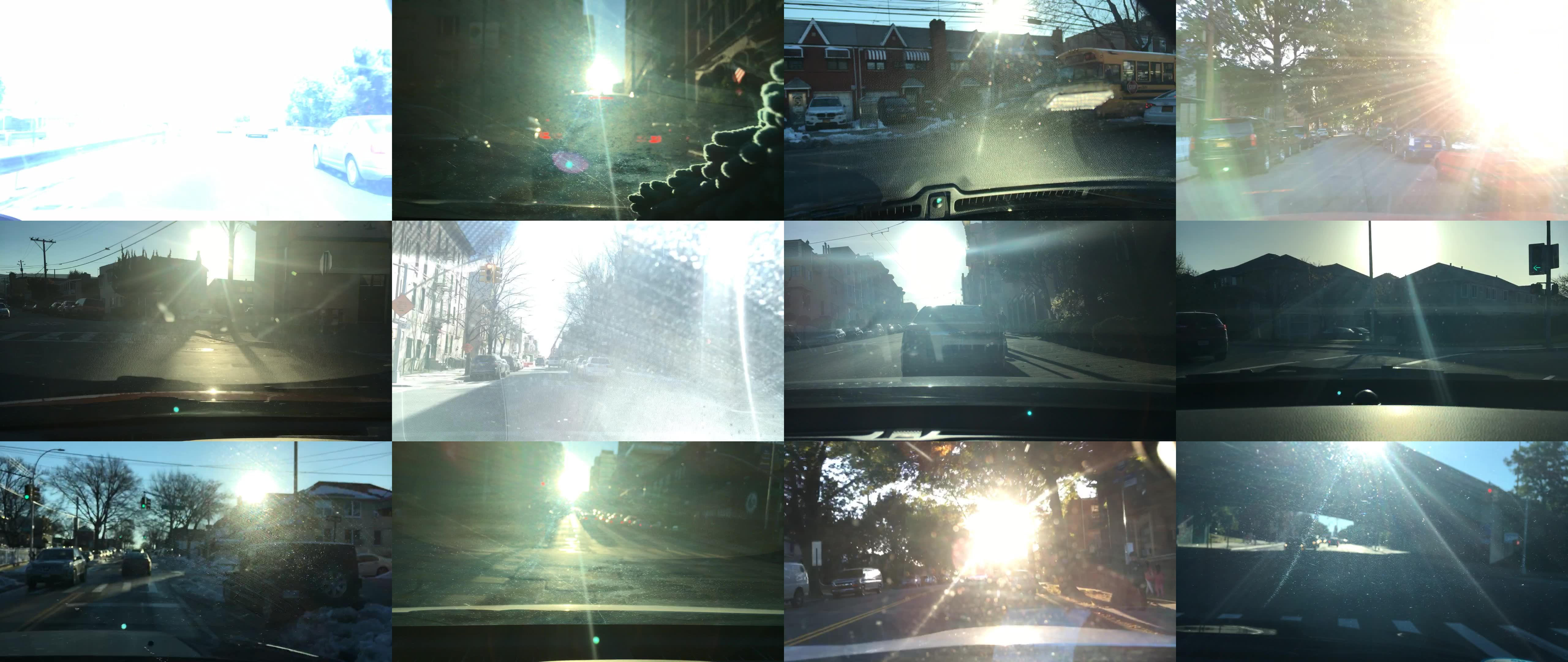}%
	\caption{Most similar images found in the BDD100k data set~\cite{yu_bdd100k_2020} with the prompt \texttt{overexposed} sorted from left to right and top to bottom.}%
	\label{fig:overexposed}%
\end{figure}

Next, we examine an example of a situation where we suspect a functional weakness of the ADS.
An overexposed image is an example of a situation where camera-based ADS could be challenged.
As a prompt, we have chosen \texttt{overexposed} (see Fig.~\ref{fig:overexposed}).
When manually inspecting the results, the first $68$ images were visibly overexposed.
The first image which was not overexposed was the $69$th image.
However, we could not identify a drop in the values of the similarity measure between the $68$th and the $69$th image.

\begin{figure}%
	\centering
	\includegraphics[width=.98\columnwidth]{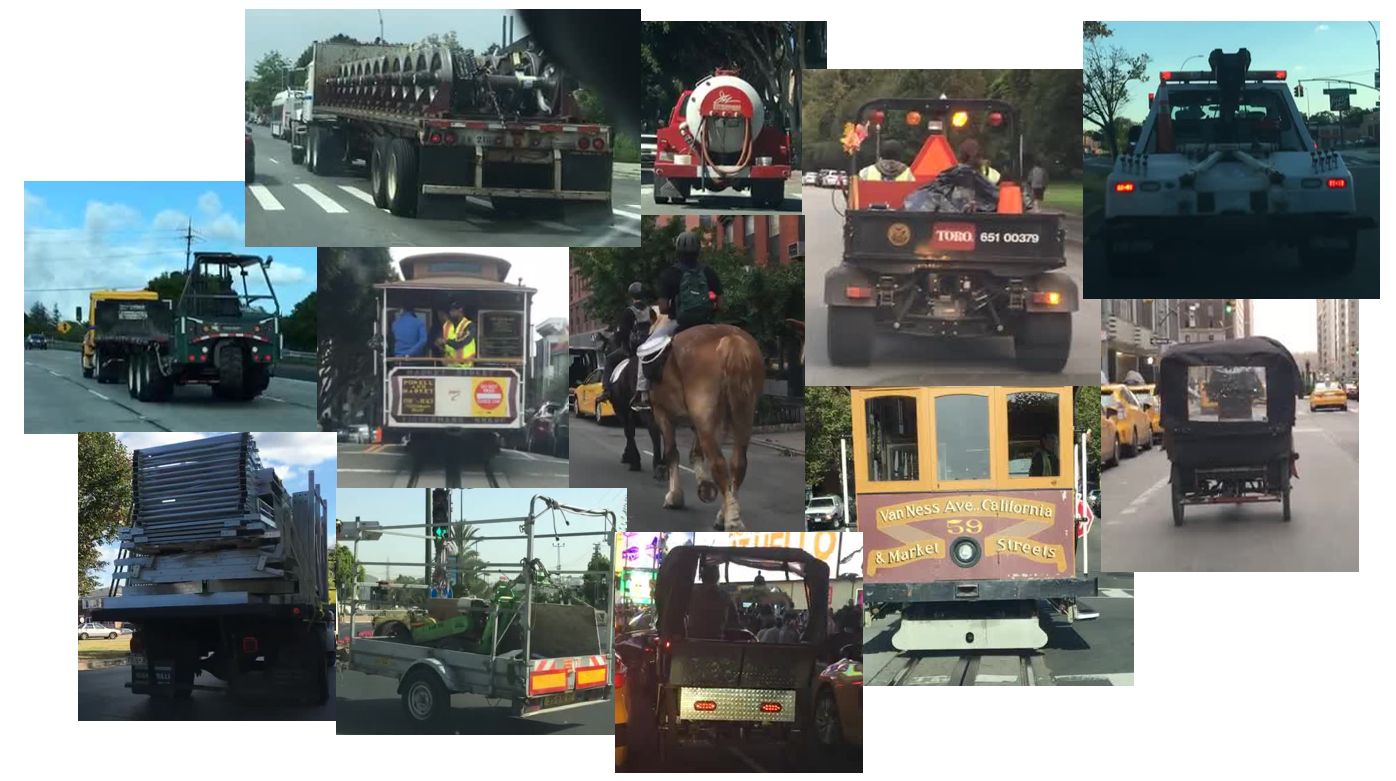}%
	\caption{Selection of images found in the BDD100k data set~\cite{yu_bdd100k_2020} with our method when starting with the word \texttt{carriage} and searching for linked words in WordNet~\cite{miller_wordnet_1995}.}%
	\label{fig:carriage_finds}%
\end{figure}

The last exemplary experiment we conducted refers to a part of an image.
As an example, we decided to choose a situation where a perceptual method showed a peculiar behavior in the real-world application.
In the considered case, inconsistencies occur in the recognition of the object type and the direction of movement of a carriage\footnote{Car Computer Confused by Carriage, (Aug. 13, 2022). [Online Video]. Available: \url{https://viralhog.com/v?t=nads5rd619}}.
To identify suitable prompts, we start with the word \texttt{carriage} and query Wordnet~\cite{miller_wordnet_1995} for linked words (see Table~\ref{tab:wordnet}).
All results were used as CLIP prompts.
A selection of the resulting images is shown in Fig.~\ref{fig:carriage_finds}.
Although no carriage was found in the data set, several atypical vehicles could be identified.

\begin{table}
	\centering
	\caption{Wordnet~\cite{miller_wordnet_1995} linkages for the word \texttt{carriage} with the definition a 'vehicle with wheels drawn by one or more horses'.}%
	\label{tab:wordnet}%
	\begin{tabularx}{\columnwidth}{lX}
		\toprule
		\textbf{Linkage  type} & \textbf{Results}\\
		\midrule
		synonym & \texttt{carriage}, \texttt{equipage}, \texttt{rig}\\
		antonym & -\\
		hypernym & \texttt{horse-drawn\_vehicle}\\
		hyponym & \texttt{barouche}, \texttt{brougham}, \texttt{buckboard}, \texttt{buggy}, \texttt{roadster}, \texttt{cab}, \texttt{cabriolet}, \texttt{caroche}, \texttt{chaise}, \texttt{shay}, \texttt{chariot}, \texttt{clarence}, \texttt{coach}, \texttt{four-in-hand}, \texttt{coach-and-four}, \texttt{droshky}, \texttt{drosky}, \texttt{gharry}, \texttt{gig}, \texttt{hackney}, \texttt{hackney\_carriage}, \texttt{hackney\_coach}, \texttt{hansom}, \texttt{hansom\_cab}, \texttt{landau}, \texttt{post\_chaise}, \texttt{stanhope}, \texttt{surrey}, \texttt{trap}, \texttt{troika}\\
		meronymy & \texttt{axletree}, \texttt{rumble}\\
		holonymy & -\\
		\bottomrule
	\end{tabularx}
\end{table}

\begin{figure}%
	\centering
	\includegraphics[width=.98\columnwidth]{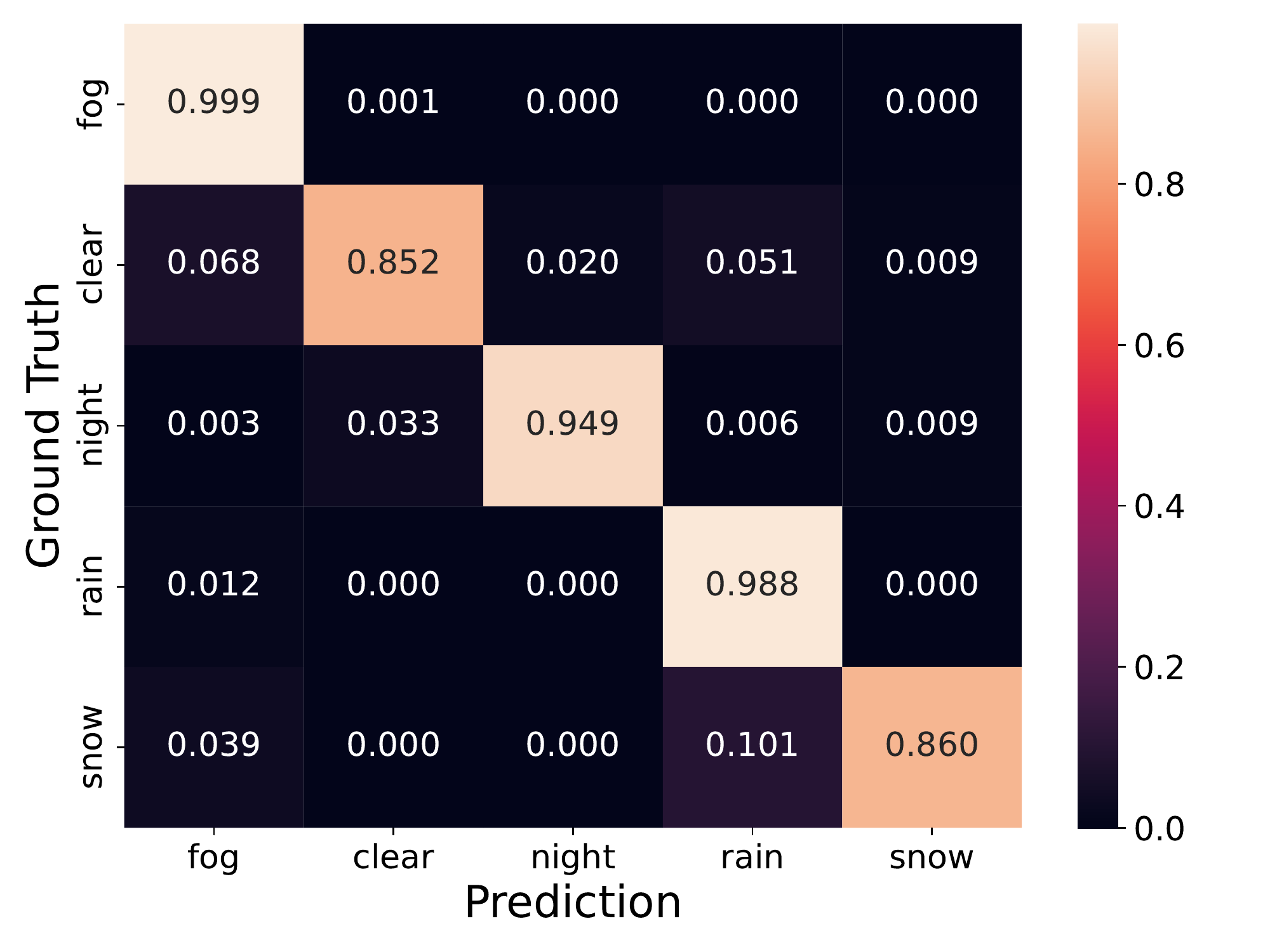}%
	\caption{Column-wise normalized confusion matrix of the condition classification with the presented method.}%
	\label{fig:confusion}%
\end{figure}

To quantitatively substantiate the operation of the method on automotive data, we use the ACDC data set~\cite{sakaridis_acdc_2021} and perform a classification experiment.
The data set was recorded at diverse conditions: fog, nighttime, rain, and snow.
For every image exists a reference image which was recorded from roughly the same perspective at daytime with clear weather.
This results in five different classes: $C = \{\texttt{clear}, \texttt{fog}, \texttt{night}, \texttt{rain}, \texttt{snow}\}$.
With CLIP, an embedding for every image is calculated.
Based on the textual description of each condition, we select a prompt for it (see Table~\ref{tab:acdc}).
\begin{table}
	\centering
	\caption{Image condition classes in the ACDC data set~\cite{sakaridis_acdc_2021}.}%
	\label{tab:acdc}%
	\begin{tabular}{lcc}
		\toprule
		\textbf{Condition} & \textbf{Image count} & \textbf{Prompt}\\
		\midrule
		clear & 4006 & \texttt{clear}\\
		fog & 1000 & \texttt{fog}\\
		nighttime & 1006 & \texttt{night}\\
		rain & 1000 & \texttt{rain}\\
		snow & 1000 & \texttt{snow}\\
		\bottomrule
	\end{tabular}
\end{table}
For every image $i$ the similarity between their CLIP embedding and every condition embedding $\vec{v}_{\text{prompt},c}, c \in C$ is calculated:
\begin{equation}
	s_{i,c} = \text{sim}_\text{cos}(\vec{v}_{\text{img},i}, \vec{v}_{\text{prompt},c}).
\end{equation}
Subsequently, to predict the condition for every image, the condition with the highest similarity score for this image is selected:
\begin{equation}
	c_i = \underset{c \in C}{\text{argmax}}(s_{i,c}) = \underset{c \in C}{\text{argmax}}(\text{sim}_\text{cos}(\vec{v}_{\text{img},i}, \vec{v}_{\text{prompt},c})).
\end{equation}
This classification $c_i$ is compared to the ground truth condition with a confusion matrix in Fig.~\ref{fig:confusion}.
Additionally, we calculated the macro F1 score~\cite{chinchor_muc-4_1992}, with the best value at 1 and the worst value at 0, for the predictions: $0.89$.

\begin{figure*}%
	\centering
	\includegraphics[width=.98\textwidth]{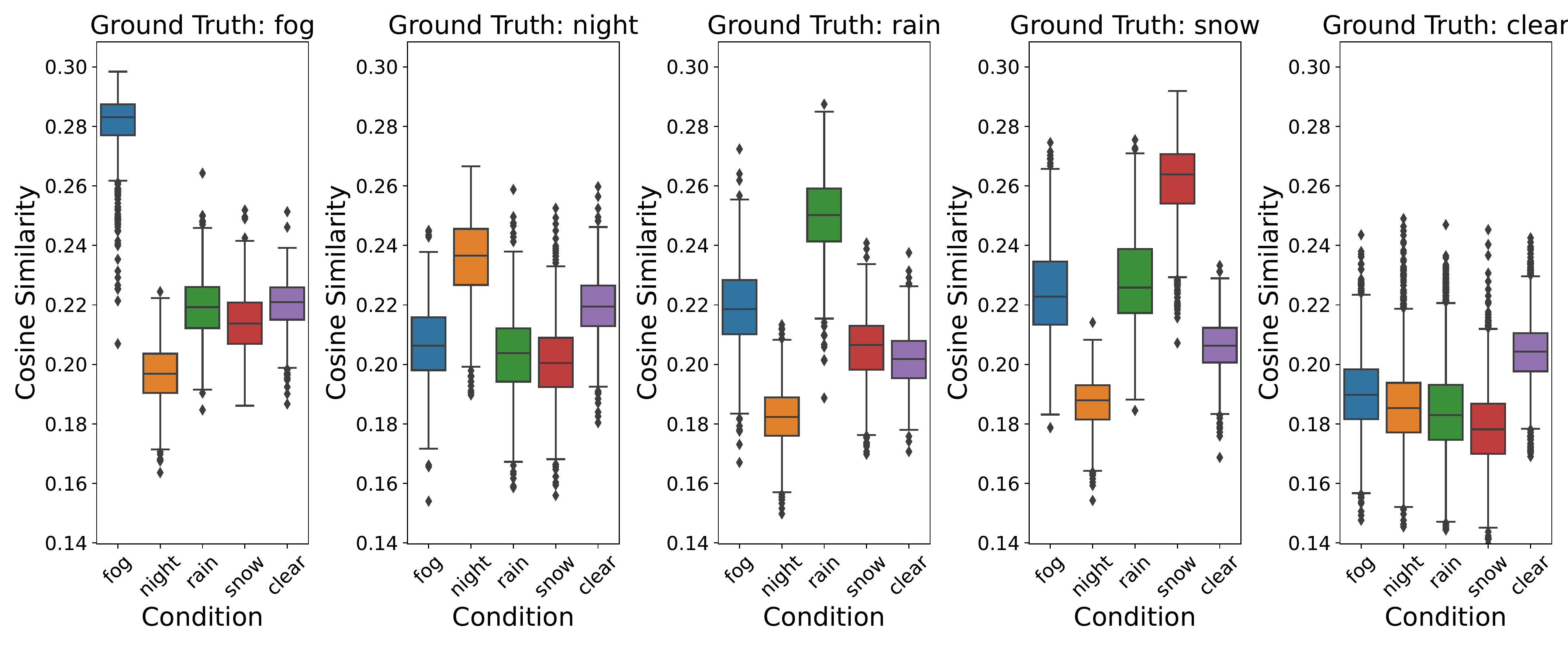}%
	\caption{Cosine similarity values between the images and condition prompts grouped by the ground truth conditions.}
	\label{fig:clipvalues}%
\end{figure*}

Fig.~\ref{fig:clipvalues} shows the cosine similarity values broken down for each condition.
The behavior already observed in the confusion matrix, that the condition \texttt{clear} cannot be well distinguished from the rest of the conditions, is also evident here.
An explanation for this could be the terse and ambiguous prompt \texttt{clear}.

\begin{figure}
	\centering
	\begin{subfigure}[t]{\columnwidth}
		\centering
		\includegraphics[width=0.9\columnwidth]{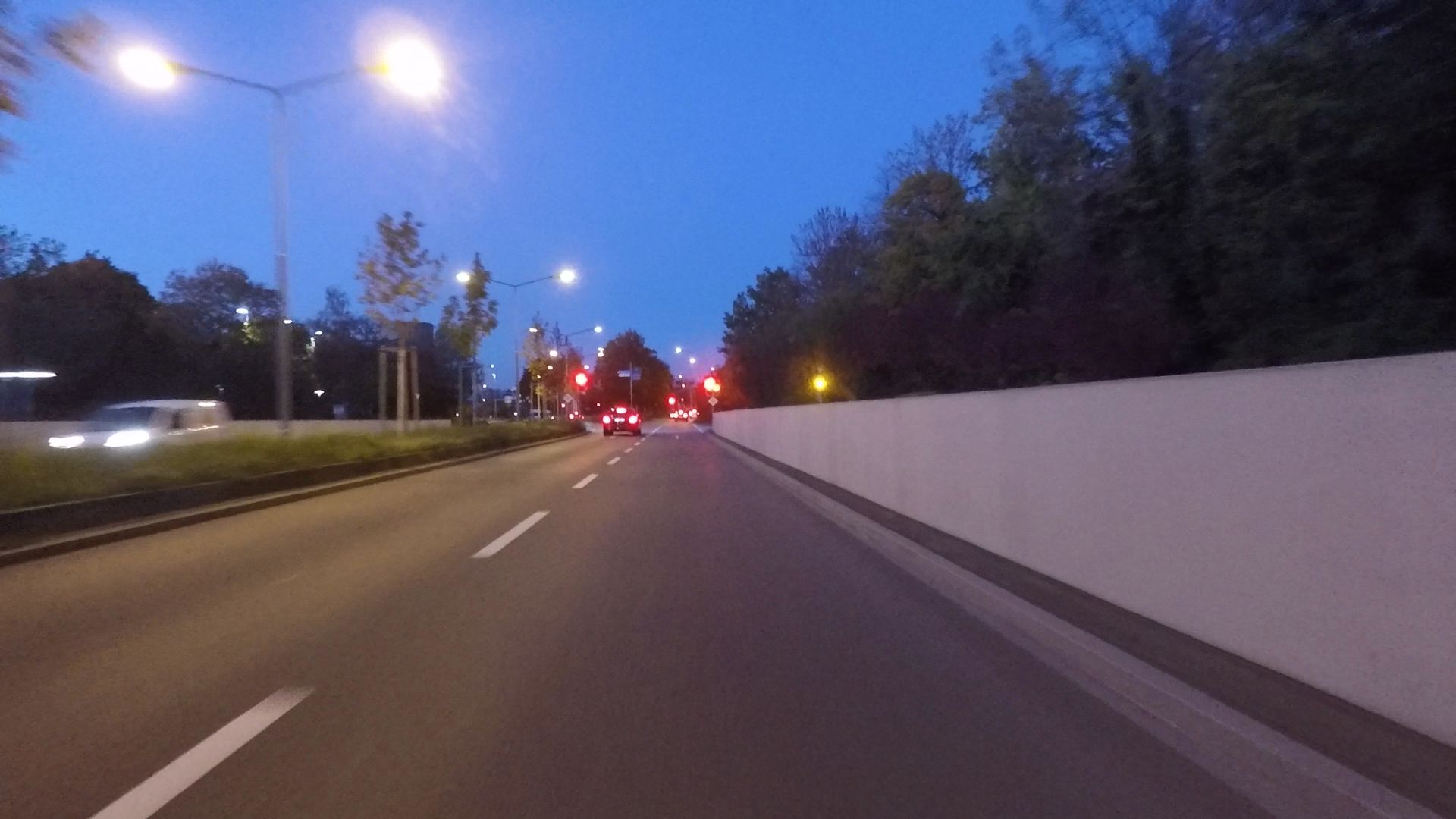}
		\caption{Ground truth: \texttt{clear} Prediction: \texttt{night}.}
		\label{fig:examples:a}
	\end{subfigure}
	\centering
	\begin{subfigure}[t]{\columnwidth}
		\centering
		\includegraphics[width=0.9\columnwidth]{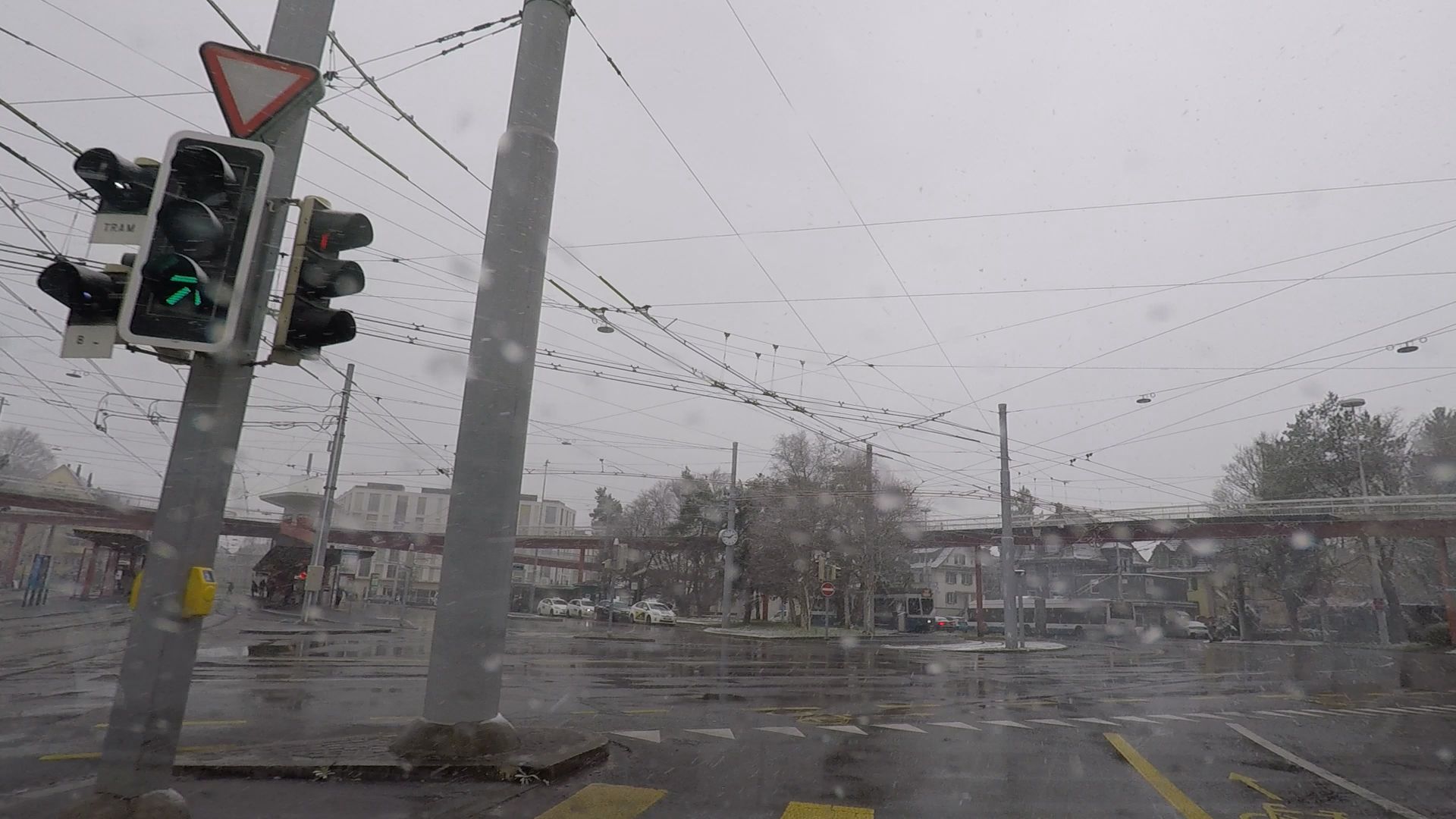}
		\caption{Ground truth: \texttt{snow} Prediction: \texttt{rain}.}
		\label{fig:examples:b}
	\end{subfigure}
	\caption{Examples of misclassifications of the conditions in the ACDC data set \cite{sakaridis_acdc_2021}.}%
	\label{fig:examples}
\end{figure}

That the classification task itself can also be ambiguous is illustrated by two examples in Fig.~\ref{fig:examples}.
It is unclear exactly where the line between day and night is drawn (see Fig.~\ref{fig:examples:a}).
In addition, it is also difficult to distinguish melted snow on the windshield from an exclusive rain situation (see Fig.~\ref{fig:examples:b}).

During our investigations, three limitations of the presented method became apparent.
Firstly, there is no way to ensure completeness when something is searched in the data set.
For example, there might be a carriage present in the BDD100k data set, but the method was unable to find it.
Secondly, false positive results also occur.
For example, rain situations could be found which are actually melted snow on the windshield.
Lastly, the method is not independent of the images itself.
Therefore, an aspect which is not visible in an image or which is not detected by CLIP could not be found by this method.
However, the method can be used for diverse data handling tasks and is not limited to predefined image aspects.

This flexibility allows a generalized use even for initially unknown tasks.
After an initial parallelizable vector calculation of the data set, only the vectors of the prompts have to be calculated at runtime.
We measured the runtime of the CLIP embedding calculation on two graphics cards (see Tab.~\ref{tab:runtime}).
As text prompts, we used three examples: \texttt{carriage}, \texttt{car} and \texttt{horse with carriage}.
The CLIP embedding for every prompt was calculated $\SI{10000}{}$ times and averaged.
The CLIP image embedding runtime was calculated with a random BDD100k image and also averaged over $\SI{10000}{}$ executions.
\begin{table}
	\centering
	\caption{Runtime of one CLIP embedding calculation averaged over $\SI{40000}{}$ executions for the text prompt and $\SI{10000}{}$ executions for the image prompt.}%
	\label{tab:runtime}%
	\begin{tabular}{ccc}
		\toprule
		\textbf{Graphics card} & \makecell{\textbf{Runtime} \\ \textbf{text prompt}} & \makecell{\textbf{Runtime} \\ \textbf{image prompt}}\\
		\midrule
		NVIDIA GeForce RTX 3090 & $\approx\SI{10}{\milli\second}$ & $\approx\SI{18}{\milli\second}$\\
		NVIDIA RTX A6000 & $\approx\SI{7}{\milli\second}$ & $\approx\SI{14}{\milli\second}$\\
		\bottomrule
	\end{tabular}
\end{table}
The cosine similarity is then calculated exclusively on the vectors, and the actual images no longer need to be considered.
This enables short runtime and scalability to large data sets.
The flexible and easy-to-understand prompt interface guarantees user-friendly use, even without in-depth expert knowledge.

\section{CONCLUSIONS}
\label{sec:conclusion}
In this contribution, we investigate the current challenge of handling image data in the automotive domain.
Due to the increasing amount of data, required for the development and test of automated driving systems, data handling has become an increasingly difficult task.
We discussed existing approaches such as metadata-based and text-based descriptions for data handling.
However, current approaches lack in automatability, generalizability and are limited in their expressiveness.

In this paper, we incorporate CLIP, a current state-of-the-art text and image embedding neural network, for handling large image data sets and guide through the application in the automotive domain.
We show how CLIP can be used with a human understandable text-based description for searching images in large data sets.
By using the lexical database WordNet, we can make full use of related terms for the text-based description.
We also investigate the search for similar images based on a sample image.
Our evaluation showed the generalizability and automatability of our proposed method for handling large data sets in the automotive domain.

However, these findings provide additional information about current limitations of the presented method: ensuring completeness for the search, occurring false positives and the dependence of the image data itself.
If the identified constraints are considered and the method is applied accordingly, the presented method provides a powerful tool for the data set handling task in the development of ADS.

Besides focusing on tackling those limitations, another interesting question for future research is the generation of new data.
In addition to browsing existing real-world data, one could utilize generative neural networks such as diffusion models~\cite{rombach_high-resolution_2022} for the generation of images.
Based on our presented method, we assume that the text prompts could be used in the text-to-image process of these diffusion models.


\section*{ACKNOWLEDGMENT}
The research leading to these results is funded by the German Federal Ministry for Economic Affairs and Energy within the project “KI Data Tooling – Methoden und Werkzeuge für das Generieren und Veredeln von \mbox{Trainings-}, Validierungs- und Absicherungsdaten für KI-Funktionen autonomer Fahrzeuge" (19A20001J). The authors would like to thank the consortium for the successful cooperation.

\bibliographystyle{IEEEtran}
\bibliography{IEEEabrv,references}

\end{document}